\documentclass[final]{cvpr}
\pdfoutput=1
\usepackage{enumitem}
\usepackage{amsmath}
\usepackage{times}
\usepackage{graphicx}
\usepackage{bm}
\usepackage{amsfonts}
\usepackage{multirow}
\usepackage{color}
\usepackage{booktabs}
\usepackage{ifthen}
\usepackage{gensymb}
\usepackage{mathtools}
\usepackage{hyperref}
\usepackage[capitalize]{cleveref}

\usepackage[normalem]{ulem} 
\usepackage{array} 
\usepackage{multirow}
\newcolumntype{C}[1]{>{\centering\arraybackslash}p{#1}}

\definecolor{gray}{rgb}{0.5,0.5,0.5}
\definecolor{green}{rgb}{0, 0.6, 0}
\definecolor{orange}{rgb}{1, 0.5, 0}
\definecolor{mahogany}{rgb}{0.75, 0.25, 0.0}
\definecolor{purple}{rgb}{0.6, 0, 0.6}
\definecolor{darkgreen}{rgb}{0, 0.3, 0}
\definecolor{orange}{rgb}{1, 0.5, 0.}

\renewcommand{\ie}{i.e.,}

\newcommand{\figname}{Figure}

\newcommand{\eqname}{Eq.}

\DeclareMathOperator*{\argmin}{arg\,min}

\def\cvprPaperID{666} 
\def\confName{CVPR}
\def\confYear{2022}

\begin{document}
\title{NeRF-In: Free-Form NeRF Inpainting with RGB-D Priors}

\author{Hao-Kang Liu\footnotemark[1]\\
National Taiwan University\\
\and
I-Chao Shen\footnotemark[1]\\
The University of Tokyo\\
\and
Bing-Yu Chen\\
National Taiwan University\\
}

\twocolumn[{%
\renewcommand\twocolumn[1][]{#1}%
\maketitle
\vspace{-1cm}
\begin{center}
    \centering
    \includegraphics[width=\linewidth]{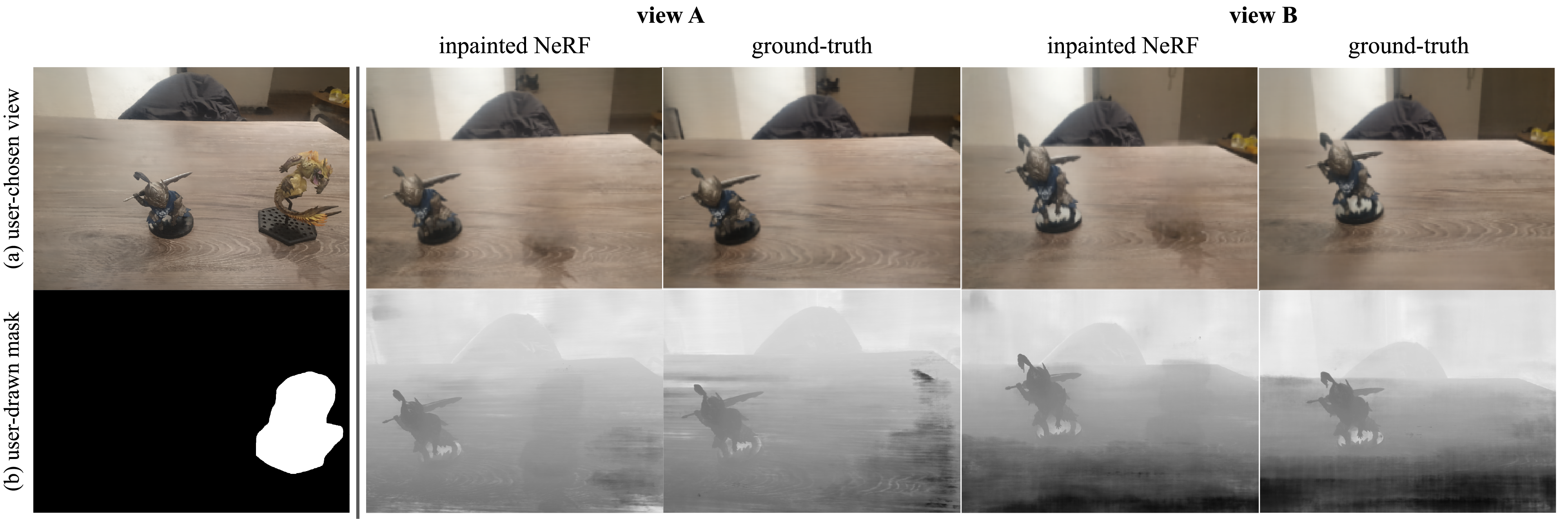}
    \captionof{figure}{    
    Given a pre-trained NeRF model, the user can (a) choose a view and (b) draw a mask to specify the unwanted object in the 3D scene.
Our framework optimized the NeRF model based on user-provided mask and remove the unwanted object in the mask region. 
The optimized NeRF generated by our framework synthesize inpainted result resembles ground truth result in different views.
}
\label{fig:teaser}
\end{center}%
}]
    
\begin{abstract}
Though Neural Radiance Field (NeRF) demonstrates compelling novel view synthesis results, it is still unintuitive to edit a pre-trained NeRF because the neural network's parameters and the scene geometry/appearance are often not explicitly associated.
In this paper, we introduce the first framework that enables users to remove unwanted objects or retouch undesired regions in a 3D scene represented by \textbf{a pre-trained NeRF} without any category-specific data and training.
The user first draws a free-form mask to specify a region containing unwanted objects over a rendered view from the pre-trained NeRF.
Our framework first transfers the user-provided mask to other rendered views and estimates guiding color and depth images within these transferred masked regions.
Next, we formulate an optimization problem that jointly inpaints the image content in all masked regions across multiple views by updating the NeRF model's parameters.
We demonstrate our framework on diverse scenes and show it obtained visual plausible and structurally consistent results across multiple views using shorter time and less user manual efforts.
\end{abstract}

\footnotetext[1]{Authors contributed equally to this work.}

\section{Introduction}

\label{sec:intro}
Recent advancements in neural rendering, such as Neural Radiance Fields (NeRF)~\cite{mildenhall2020nerf} has emerged as a powerful representation for the task of novel view synthesis, where the goal is to render unseen viewpoints of a scene from a given set of input images.
NeRF encodes the volumetric density and color of a scene within the weights of a coordinate-based multi-layer perceptron.
Several follow-up works extend original NeRF to handle different tasks, such as pose estimation~\cite{lin2021barf, wang2021nerf}, 3D-aware image synthesis~\cite{schwarz2020graf, niemeyer2021giraffe, chan2021pi},  deformable 3D reconstruction~\cite{Pumarola20arxiv_D_NeRF, liu2021neuralactor, park2021nerfies}, and modeling dynamic scenes~\cite{xian2021space, guo2021ad, Gao-freeviewvideo}.

Though NeRF achieves great performance on photo-realistic scene reconstruction and novel view synthesis, there remain enormous challenges in editing the geometries and appearances of a scene represented by a pre-trained NeRF model.
Unlike traditional image editing, a user needs to transfer his/her edits on a rendered view to the NeRF model to edit the whole scene, thus introducing multiple challenges.
First, it is unclear where the edited regions appear on other rendered views.
Second, because millions of parameters are used in a pre-trained NeRF model, it is unclear which parameters control the different aspects of the rendered shape and how to change the parameters according to the sparse local user input.
Previous works~\cite{liu2021editing} enable users to perform color and shape editing on a category-level NeRF.
However, these methods require additional category-specific training and data to support the desired editings.

In this paper, we focus on the NeRF inpainting problem, \ie~removing unwanted objects in a 3D scene represented by a pre-trained NeRF.
Although we can ask a user to provide a mask and the inpainted image for each rendered view and use these images to train a new NeRF, there are several disadvantages.
First, it is labor-intensive to provide masks for many rendered views.
Second, there will be visual inconsistency across different inpainted views introduced by separate inpainting.


To address these issues, we propose a framework to help users easily remove unwanted objects by updating a pre-trained NeRF model.
Given a pre-trained NeRF, the user first draws a mask over a rendered view.
Given the user-drawn mask, our framework first rendered a couple of views sampled from a pre-set trajectory.
Next, we transfer the user-drawn masks to these sampled views using existing video object segmentation method~\cite{cheng2021rethinking}. 
Our framework then generates (i) guiding color image regions using \cite{cao2021learning} and (ii) guiding depth images using Bilateral Solver~\cite{BarronPoole2016} within these masked regions.
Noted that our framework can use any existing methods for generating guiding color and depth images.
Finally, we formulate an optimization problem that jointly inpaints the image content within the all transferred masked regions with respect to the guiding color and depth images.

We demonstrate our framework on several scenes represented by pre-trained NeRFs in LLFF dataset.
We show that our framework generates visually plausible and consistent results.
Furthermore, we also demonstrate our experiments on the custom dataset to show the correctness between inpainted results and ground truth results.


\section{Related Work}
\label{sec:related}
\subsection{Novel view synthesis}
Constructing novel views of a scene captured by multiple images is a long standing problem in computer graphics and computer vision.
Traditional methods use structure-from-motion~\cite{hartley2003multiple} and bundle adjustment~\cite{triggs1999bundle} to reconstruct explict point cloud structure and recover camera parameters.
Other methods synthesize novel views by interpolating within a 4D light fields~\cite{gortler1996lumigraph,levoy1996light} and by compositing the warped layers in the multiplane image representations (MPI)~\cite{tucker2020single, zhou2018stereo}.
Recently, the coordinate-based neural representations have shown significant promise as an alternative to discrete, grid representations for scene representations. 
Neural Radiance Fields (NeRF)~\cite{mildenhall2020nerf} use a multi-layer perceptron (MLP) and positional encoding to model a radiance field at an unprecedented level of fidelity.
However, NeRF works on a large number of input images and requires lengthy training time.
Many works attempt to reduce the number of images NeRF requires by introducing depth-supervised loss~\cite{deng2021depth,roessle2021dense} and category-specific priors~\cite{yu2021pixelnerf}. 
Meanwhile, previous works try to reduce the training time by optimizing voxel grids of features~\cite{sun2021direct,yu2021plenoxels} and factoring the radiance field~\cite{chen2022tensorf}.

These recent advances greatly improve the practical use of NeRF.
However, it is still unintuitive how a user can edit a pre-trained NeRF model.
The main reason is because the neural network of a NeRF model has millions of parameters.
Which parameters control the different aspects of the rendered shape and how to change the parameters to achieve desired edtis are still unknown.
Previous works enable users to select certain object~\cite{ren-cvpr2022-nvos}, edit a NeRF model using strokes~\cite{liu2021editing}, natural language~\cite{wang2021clip}, and by manipulating 3D model~\cite{yuan2022nerf} directly.
However, these methods require to learn additional category-level conditional radiance fields or segmentation network to facilitate such edits.
Unlike these methods, our framework did not requires the user to prepare any additional category-specific training data and training procedure for removing unwanted objects in a pre-trained NeRF model.

\subsection{Image inpainting}
In recent years, two broad approaches to image inpainting exist. 
Patch-based method~\cite{efros1999texture,simakov2008summarizing,barnes2009patchmatch} fill the holes by searching for patches with similar low-level image features such as rgb values. 
The search space can be the non-hole region of the input image or from other reference images.
The inpainted results are obtained by a global optimization after the relevant patches are retrieved.
These methods often fail to handle large holes where the color and texture variance is high.
Meanwhile, these methods often cannot make semantically aware patch selections.
Deep learning-based methods often predict the pixel values inside masks directly in a semantic-aware fashion.
Thus they can synthesize more visually plausible contents especially for images like faces~\cite{li2017generative,yeh2017semantic}, objects~\cite{pathak2016context} and natural scenes~\cite{iizuka2017globally}.
However, these methods often focus on regular masks only.
To handle irregular masks, partial convolution~\cite{liu2018image} is proposed where the convolution is masked and re-normalized to utilize valid pixels only.
Yu~\etal~\cite{yu2018generative} uses GAN mechanism to maintain local and global consistency in the final results.
Nazeri~\etal~\cite{nazeri2019edgeconnect} focus on improving the image structure in the inpainting results by conditioning their image inpainting network on edges in the masked regions. MST inpainting~\cite{cao2021learning} and ZITS~\cite{dong2022incremental} further consider both edge and line structure to synthesize more reasonable results.
In this work, we use MST inpainting network~\cite{cao2021learning} to obtain the guided inpainted result because of its superior performance on inpainting images while preserving structures.
Our framework can replace MST inpainting with other inpainting methods since we only used the inpainted results as a guiding signal for our optimization problem.

\begin{figure*}
    \centering
    \includegraphics[width=\linewidth]{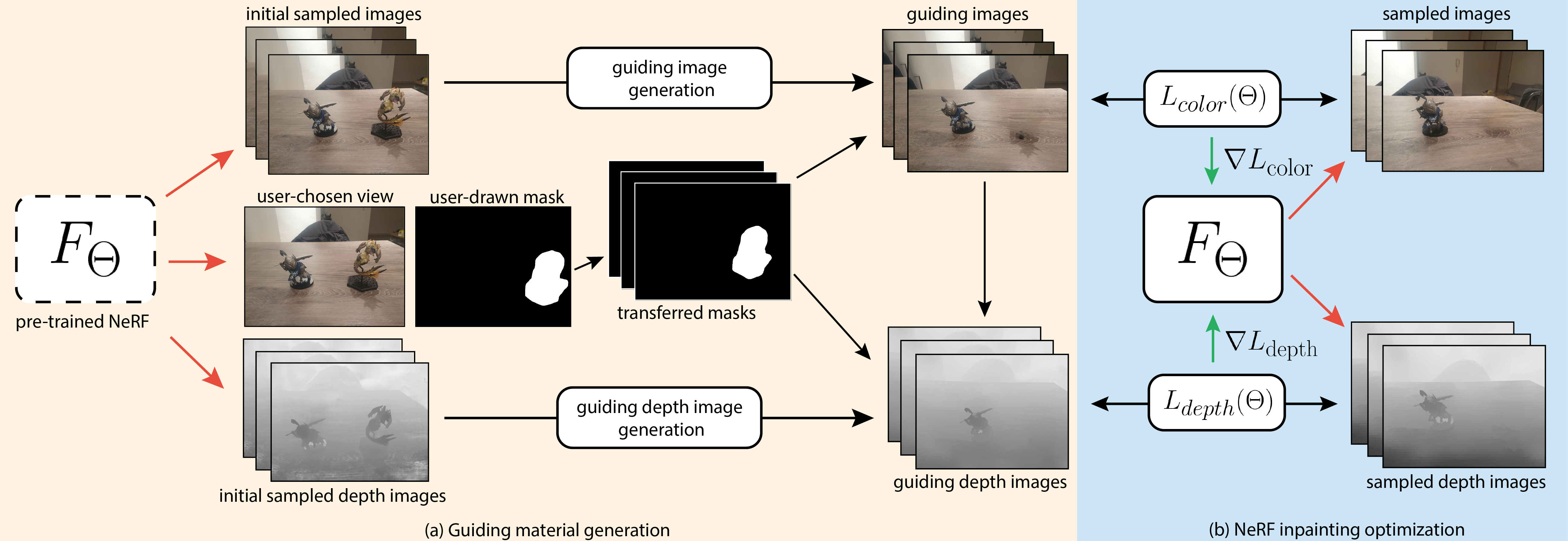}
    \caption{
    (a) Given a pre-trained NeRF $F_{\Theta}$, an user specifies the unwanted region on an user-chosen view with a user-drawn mask.
    Our framework sampled initial images and initial depth images and generate both guiding images and guiding depth images.
    (b) Our framework update $\Theta$ by optimizing both color-guiding loss ($L_{\text{color}}$) and depth-guiding loss ($L_{\text{depth}}$).
    (\textcolor[rgb]{0.9,0.298,0.23}{$\boldsymbol{\rightarrow}$} denotes render a view from a NeRF model and \textcolor[rgb]{0.15,0.68,0.38}{$\boldsymbol{\rightarrow}$} denotes updating $\Theta$ by optimizing losses.)
    }
    \label{fig:overview}
\end{figure*}
\section{Method}
In this section, we first summarize the mechanism of NeRF~\cite{mildenhall2020nerf} and formulate our problem setting.
\subsection{Preliminaries: NeRF}
NeRF is a continuous volumetric radiance field $F_{\Theta}:(\mathbf{x},\mathbf{d}) \rightarrow (\mathbf{c}, \sigma)$ represented by a MLP network with $\Theta$ as its weights.
$F_{\Theta}: (\mathbf{x},\mathbf{d}) \rightarrow (\mathbf{c}, \sigma)$ takes a 3D position $\mathbf{x} = \{x,\ y,\ z\}$ and 2D viewing direction $\mathbf{d} = \{\theta,\ \phi\}$ as input and outputs volume density $\sigma$ and directional emitted color $\mathbf{c}$.
NeRF renders the color of each camera rays passing through the scene by computing the volume rendering intergral using numerical quadrature. 
The expected color $C(\mathbf{r})$ of camera ray $\mathbf{r}(t)=\mathbf{o}+t\mathbf{d}$ is defined as:
\begin{align}
    \hat{C}(r) &= \sum_{i=1}^{N}T(t_i)(1 - \exp{(-\sigma(t_i)\delta_i)})c(t_i), \\ \text{where} \quad T(t_i)&=\exp{(-\sum_{j=1}^{i-1}\sigma(t_j)\delta_j)}\ ,
\end{align}
where $N$ denotes the total quadrature points sampled between near plane $t_n$ and far plane $t_f$ of the camera, and $\delta_i = t_{i+1} - t_i$ is the distance between two adjacent points.  
We denote the color and desity at point $t_i$ produced by NeRF model $F_{\Theta}$ as $c(t_i)$ and $\sigma(t_i)$. 

Using the above differentiable rendering equation, we can propagte the errors and update $\Theta$ through mean square error:
\begin{align}
    \mathcal{L}_{mse} =  \sum_{\mathbf{r}\in\mathcal{R}}\|(\hat{C}^c(\mathbf{r}) - C(\mathbf{r})) \|^2_2 + \|(\hat{C}^f(\mathbf{r}) - C_i(\mathbf{r}))\|^2_2,
\end{align}
where $\mathcal{R}$ is a ray batch, $C(\mathbf{r}), \hat{C}^{c}(\mathbf{r}), \hat{C}^f(\mathbf{r})$ are the ground truth, coarse volume predicted, and fine volume predicted RGB colors for ray $r$ respectively.
For simplicity, we further define $F_{\Theta}^{\text{image}}: \mathbf{o}\rightarrow I$ as a function that takes a camera position ($\mathbf{o}$) as input, and outputs the rendered image of a pre-trained NeRF model $F_{\Theta}$.
\begin{figure}
    \centering
    \includegraphics[width=\linewidth]{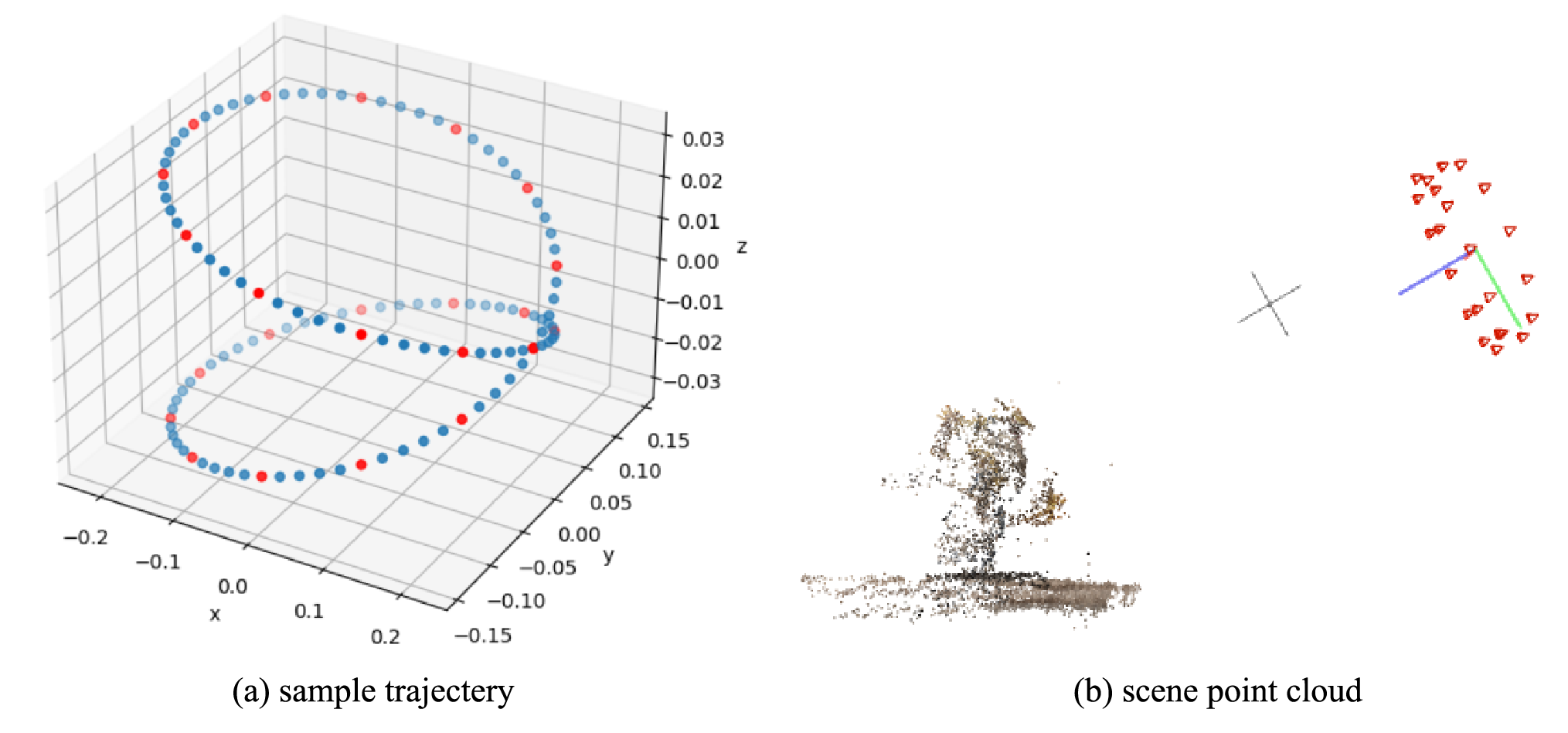}
    \caption{
    Our sampling strategy follows the trajectory in (a). 
    Noted that the target scene faces toward the ``+y'' directon as shown in (b). 
    Each blue dot represents a view we can sample, while the red dot represents the sample view used in the optimization framework. 
    Also, we use sample images to construct the point cloud for better understanding.
    We conduct all the experiments based on this setting.
    }
    \label{fig:trajectory}
\end{figure}

\subsection{Overview}
Given a pre-trained NeRF model: $F_{\Theta}$, a user can specify the unwanted region by drawing a mask $M_u$ over a user-chosen rendered view $I_u=F_{\Theta}^{\text{image}}(\mathbf{o}_u)$, where $M_u=1$ for pixel outside the masked region, and $\mathbf{o}_u$ is the user-chosen camera position.
Our goal is to obtain an updated NeRF model $F_{\tilde{\Theta}}$ such that the unwanted region masked by $M_u$ is removed in every rendered views.
As shown in \Cref{fig:overview}, our method first sample $K$ camera positions $\mathbf{O}=\{\mathbf{o}_s|s=1...K\}$ along the test-set camera trajectory used in LLFF~\cite{mildenhall2019llff} (\Cref{fig:trajectory}).
For each camera position, we rendered a rgb image $I_s$ and depth image $D_s$ using $F_{\Theta}$ and obtained all rendered views $\mathbf{I}=\{I_s|s=1 ... K\}$ and their depth images $\mathbf{D}=\{D_s |s=1...K\}$.
We will use $I_s$ and $I(\mathbf{o}_s)$ to represent the image rendered from camera position $\mathbf{o}_s$ interchangeably throughout the paper.
One can potentially remove the unwanted object specified by $M_u$ using the following naive method.
First, remove the content within the transferred masked region on each sampled rendered view.
Then, update $\Theta$ using only the image content outside all transferred masked region by optimizing the ``masked mse (mmse)'' function:
\begin{align}
L_{\text{mmse}} = \sum_{\mathbf{r}\in\mathcal{R}}\|(\hat{C}^c(\mathbf{r}) - C(\mathbf{r})) \odot M(\mathbf{r}) \|^2_2 \\ \nonumber
+ \| (\hat{C}^f(\mathbf{r}) - C_i(\mathbf{r}))\odot M(\mathbf{r})\|^2_2,
\label{eq:mmse}
\end{align}
where $M$ is the transferred mask on the same view as the sample ray $\mathbf{r}$.
However, 
because there is no explicit guidance on what image content and structure should be in the masked region, the unwanted object will remain in the result of optimizing \Cref{eq:mmse}.
To provide explicit guidance, our method takes the user-drawn mask $M_u$, sampled rendered views $\mathbf{I}$, and sampled depth images $\mathbf{D}$ as input, and outputs 
\begin{itemize}
    \item guiding user-chosen image and guiding depth image: $I_u^G$ and $D_u^G$.
    \item transferred masks: $\mathbf{M}=\{M_s | s=1...K\}$.
    \item guiding sampled images and their guiding depth images: $\mathbf{I}^{G}=\{I_s^G| s=1...K\}$ and $\mathbf{D}^{G}=\{D_s^G|s=1...K\}$.
\end{itemize} 
Finally, our method obtains updated parameters $\tilde{\Theta}$ by optimizing our nerf inpainting formulation: $\Phi(M_u, {I}_u^{G}, \mathbf{M},  \mathbf{D}_s^{G})$.

\subsection{Guiding material generation}
For each sampled rendered view $I_s$, our goal is to generate a mask $M_s$ that covers the same object as the user-drawn mask $M_u$.
We use a video object segmentation method (STCN)~\cite{cheng2021rethinking} to generate $M_s$.
With the transferred masks $M_s$, we need to generate the guiding images and guiding depth images.
The guiding image generation can be describe as 
\begin{align}
I^G_s=\rho(I_s, M_s),
\end{align}
where $I^G_s$ is the guiding image, and $\rho$ is a single image inpainting method (we used MST inpainting network~\cite{cao2021learning}).
After obtaining $I^G_s$, we can obtain the guiding depth image using
\begin{align}
D_s^G = \tau(D_s, M_s, I^G_s),
\end{align}
where $D^G_s$ is the guiding depth image, and $\tau$ is a depth image completion method (we used Fast Bilateral solver~\cite{BarronPoole2016}).
Noted that our framework can replace $\rho$ to any other single image inpainting method and $\tau$ to any other single depth image completion method.

\subsection{NeRF inpainting optimization}
We obtain the updated parameters $\tilde{\Theta}$ that removes the unwanted object in the 3D scene by optimizing:
\begin{align}
\tilde{\Theta} \coloneqq\argmin_{\Theta} \; L_{\text{color}}(\Theta) + L_{\text{depth}}(\Theta)
\label{eq:inpaint_opt}
\end{align}
where $L_{\text{color}}$ is the color-guiding loss and $L_{\text{depth}}$ is
the depth-guiding loss.

\subsubsection{Color-guiding loss}
The color-guiding loss used to is defined as 
\begin{align}
L_{\text{color}}(\Theta) = L_{\text{color}}^{\text{all}}(\Theta) + L_{\text{color}}^{\text{out}}(\Theta),
\label{eq:color}
\end{align}
where $L_{\text{color}}^{\text{all}}$ is defined on views $\mathbf{O}^{\text{all}}$, $L_{\text{color}}^{\text{out}}$ is defined on views $\mathbf{O}^{\text{out}}$, and $\mathbf{O}^{\text{all}}\cup \mathbf{O}^{\text{out}} = \{ \mathbf{O}, \mathbf{o}_u\}$.
$L_{\text{color}}^{\text{all}}$ is used to measure the color difference of the entire image (inside and outside of the mask) on the rendered view and is defined as:
\begin{align}
L_{\text{color}}^{\text{all}}(\Theta) = \sum_{\mathbf{o}\in \mathbf{O}^{\text{all}}}F^{\text{image}}_{\Theta}(\mathbf{o}) - I_o^G.
\label{eq:color_all}
\end{align}
$L_{\text{color}}^{\text{out}}$ is used to measure the color difference outside the mask on the rendered view and is defined as:
\begin{align}
L_{\text{color}}^{\text{out}}(\Theta)=  \sum_{\mathbf{o} \in \mathbf{O}^{\text{out}}} (F^{\text{image}}_{\Theta}(\mathbf{o}) - I_o^G)\odot M_o.
\label{eq:color_out}
\end{align}
In our framework, we set $\mathbf{O}^{\text{all}}=\mathbf{o}_u$ and $\mathbf{O}^{\text{out}}=\mathbf{O}$.

\subsubsection{Depth-guiding loss}
While we can obtain visual plausible inpainted color results using the color-guided loss, 
it often generates  incorrect depth, which might cause incorrect geometry and keep some unwanted objects in the scene. 
To fix these incorrect geometries, we introduce a depth-guided loss, which is defined as:
\begin{align}
    L_{\text{depth}}(\Theta) = \sum_{\mathbf{o}_s \in \mathbf{O}}  \|D^f(\mathbf{o}_s) - D^G(\mathbf{o}_s)\|^2_2
    + \|D^c(\mathbf{o}_s) - D^G(\mathbf{o}_s)\|^2_2,
\end{align}
where $D^f(\mathbf{o}_s)$ is the fine volume predicted depth image, $D^c(\mathbf{o_s})$ is the coarse volume predicted depth image, and we rendered both depth image from a sampled camera position $\mathbf{o}_s$ using $F_{\Theta}$. 
We compute the depth $D^f(\mathbf{o}_s)$ through computing the accumalation of $\sigma$ from ray batches.
\begin{figure*}[h!]
    \centering
    \includegraphics[width=\linewidth]{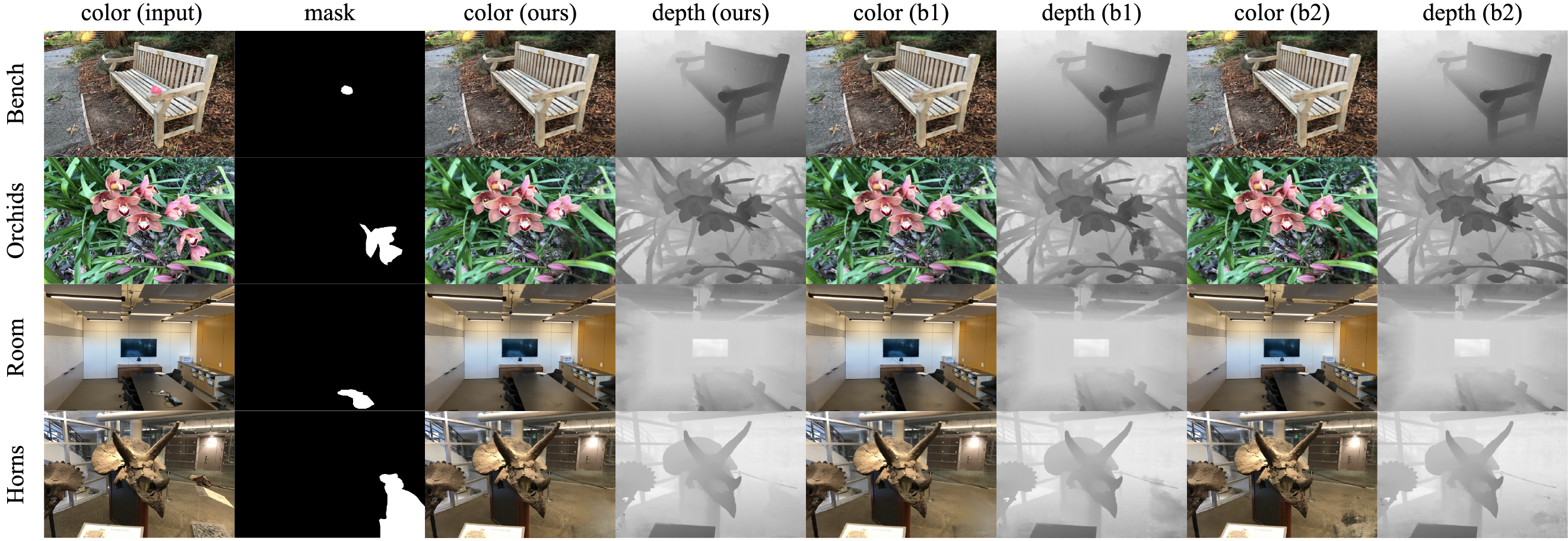}
    \caption{
    \textbf{Qualitative comparison - LLFF dataset.}
    For each scene, we show the user-chosen view image and the user-provided mask on the left.
    We then show the color image and depth image generated by different methods: our method (ours), \textit{baseline1} (b1), and \textit{baseline2} (b2). The depth map of b1 still keep depth of the unwanted object. Meanwhile, the color of b2 might cause noise or shadow on the scene(shown in horns). Our method, compared to these two baselines, have better color and correct geometry on final results.
    }
    \label{fig:llff_qual_comp}
\end{figure*}
\begin{figure*}
    \centering
    \includegraphics[width=\linewidth]{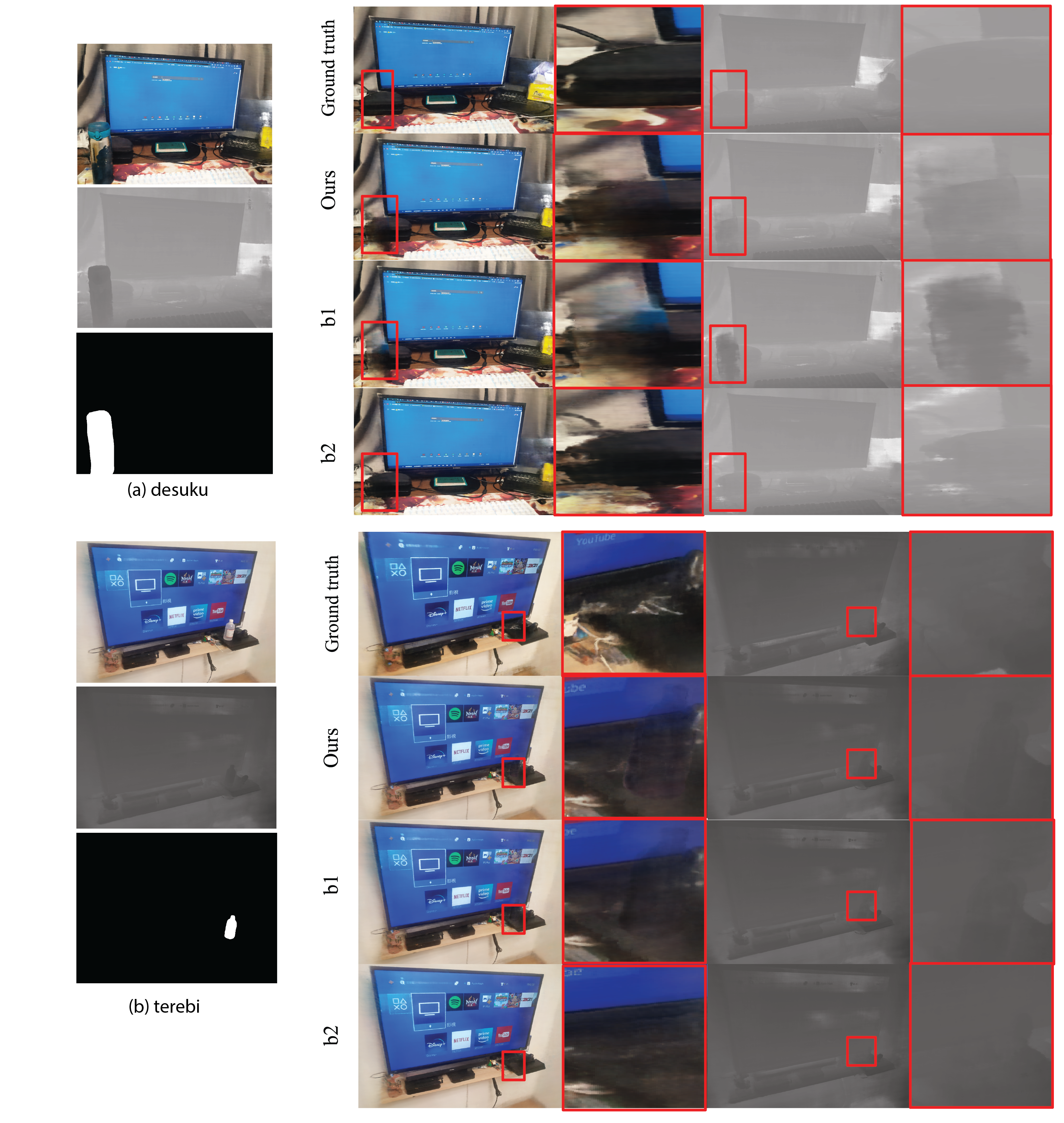}
    \caption{
    \textbf{Qualitative comparison - custom dataset.}
    For each custom secne, we demonstrate the ground truth rendered image, results generated by our framework, \textit{baseline1} (b1), and \textit{baseline2} (b2).
    Our framework generates more accurate depth maps and synthesize more fine structures compared to \textit{baseline1}.
    Compared to \textit{baseline2}, our framework synthesizes more realistic and shape results.
    }
    \label{fig:custom_result}
\end{figure*}
\begin{figure}[h!]
    \centering
    \includegraphics[width=\linewidth]{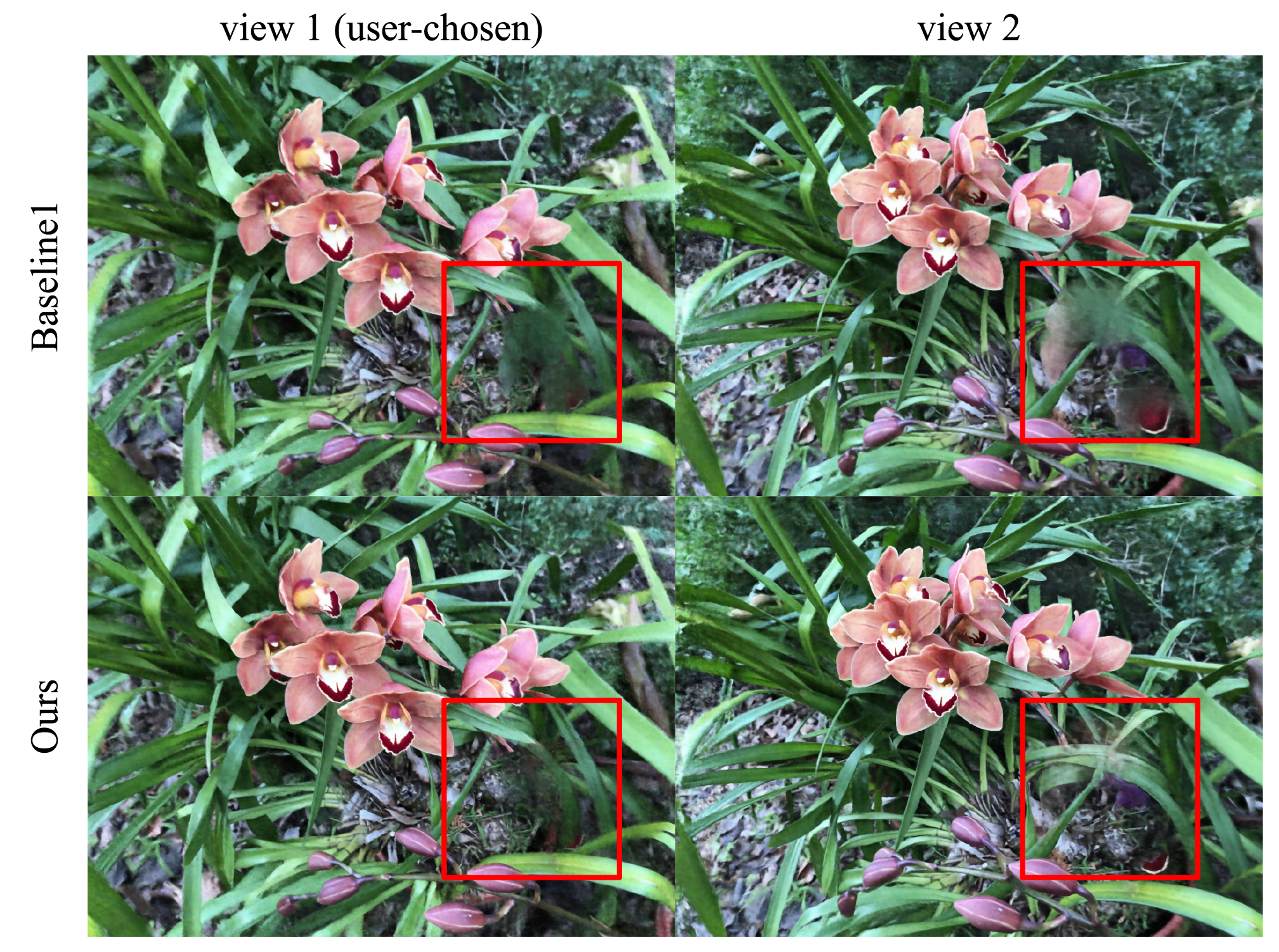}
    \caption{
    \textbf{Qualitative comparison - visual consistency.}
    The rendered views generated by \textit{baseline1} have severe visual inconsistency across different views (within the red box region).
    Meanwhile, our method synthesize visual consistent results across different views.
    }
    \label{fig:consistency}
\end{figure}
\begin{figure*}[h!]
    \centering
    \includegraphics[width=\linewidth]{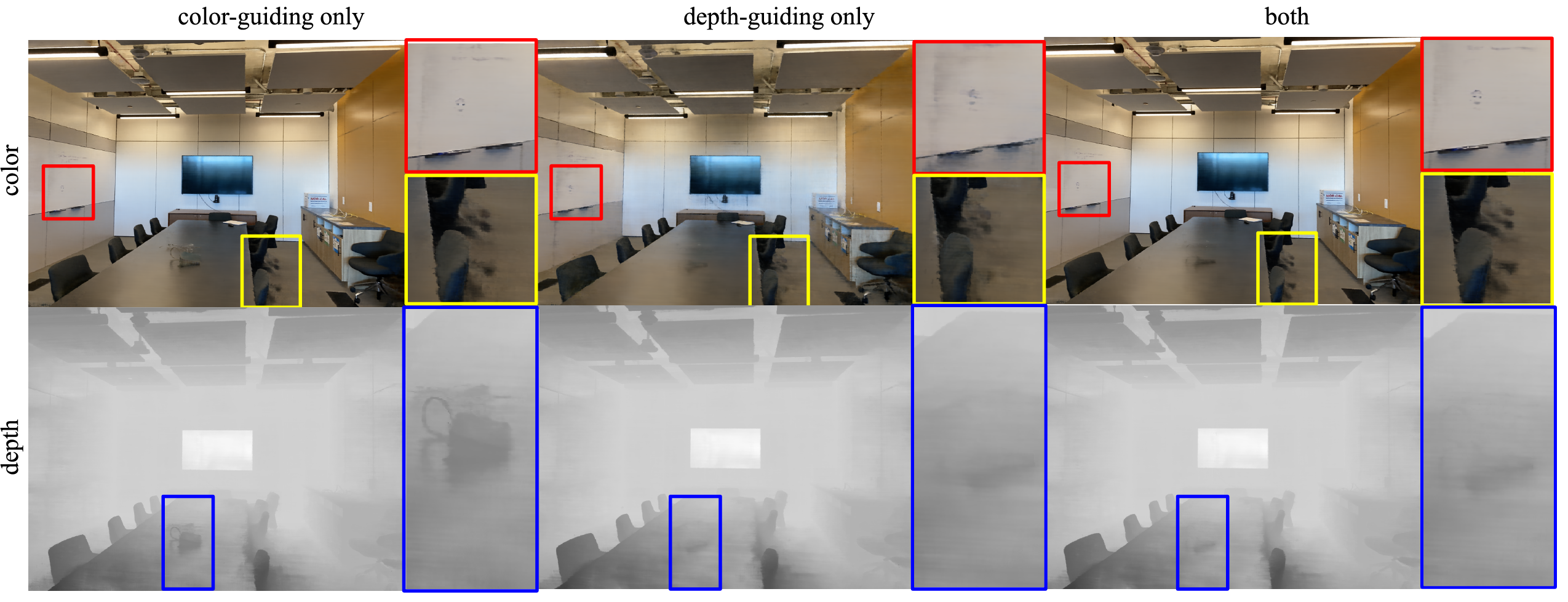}
    \caption{
    \textbf{Depth-guiding ablation result.}
    We show the optimization results using different guiding losses.
    We observed that the geometry of the unwanted object could not be removed using color-guiding term only (depth inside the blue box). 
    On the other hand, using depth-guiding term only helps to get correct geometry but introduces color noises outside the masked region (red box and yellow box). 
    Our method combines both terms to generate correct geometry (blue box) without introducing any color noises (red box and yellow box).
    }
    \label{fig:depth_ablation}
\end{figure*}

\begin{figure}[h!]
    \centering
    \includegraphics[width=\linewidth]{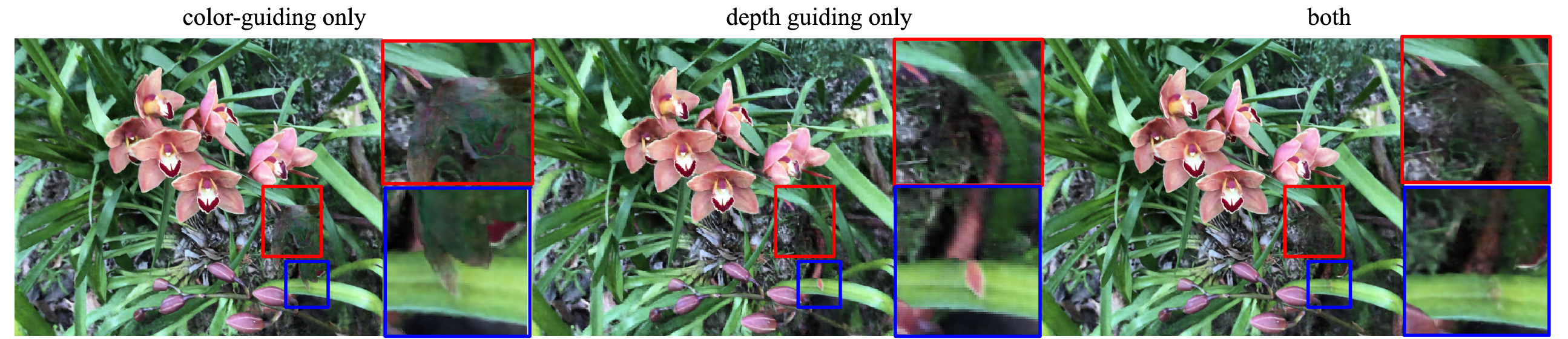}
    \caption{
    \textbf{Depth-guiding losses discussion.}
        The unwanted object in the region with high depth variations can be removed by using depth-guiding loss (red box).
        However, using depth-guiding loss only loses color information in the flat region (blue box).
        Combining both losses removes the unwanted object without losing any color information.
    }
    \label{fig:term_local_comp}
\end{figure}
\begin{figure}[h!]
    \includegraphics[width=\linewidth]{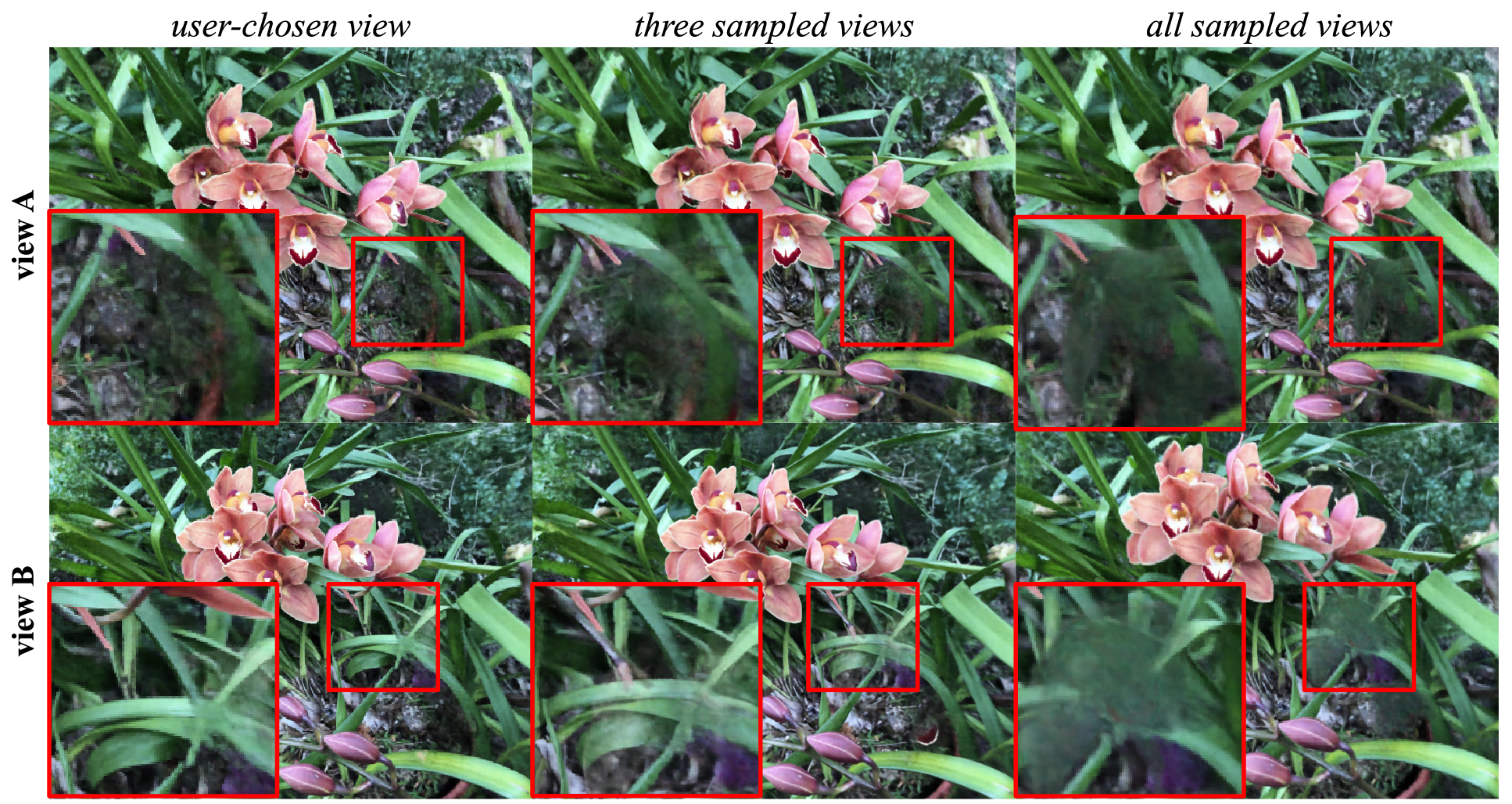}
    \caption{
    \textbf{Color-guiding ablation result.}
    We show the optimization results using different number of views to compute $L_{\text{color}}^{\text{all}}$.
    The visual inconsistency becomes larger in the masked region (red box) when we increase the number of views used to compute $L_{\text{color}}^{\text{all}}$.
    }
    \label{fig:k_setting}
\end{figure}

\section{Experiments and evaluations}
In this section, we show qualitative results on LLFF~\cite{mildenhall2019llff} dataset and our custom dataset, followed by ablation studies.
\subsection{Implementation detail}
We implement our framework in PyTorch~\cite{pytorch} and Python $3.9$. 
We test our framework on a machine with Intel i7-7800X and a GTX-1080 graphics card to train our models.
For each scene, we first train a model initialized to random weights and optimize it for $200,000$ steps with a batch size of $4,096$ using Adam~\cite{kingma2014adam}, which takes about $18$ to $20$ hours.
The sample points used in fine and coarse model are $128$ and $64$, respectively.
To inpaint each scene, we optimize \eqname~\ref{eq:inpaint_opt} for $50,000$ steps which takes about five hours in total.
\subsection{Evaluation}
\subsubsection{Datasets}
\label{sec:custom_dataset}
To verify our framework's performance, we create a custom dataset that contains three custom scenes: \textbf{figyua}, \textbf{desuku}, and \textbf{terebi}.
The purpose is to obtain the ground truth results of NeRF inpainting.
For each custom scene, we collect a pair of photo set, \ie~(\textbf{original} and \textbf{removed}). 
For \textbf{original} set, we keep all the objects in the scene and take photos from $24$ camera positions.
For \textbf{removed} set, we remove one object in the scene manually and take photos from the same $24$ camera positions.
The \textbf{original} sets is used as the input of our framework, and the \textbf{removed} set is used as the ground truth of the results after the inpainting optimization. 

\subsubsection{Experiment setup}
As we are the first to free-form inpainting on NeRF, we propose two baseline methods for comparisons: 
\begin{description}[style=unboxed,leftmargin=0cm]
\item[\textit{baseline1}: per-view color updating]
We update the pre-trained NeRF model $F_{\Theta}$ with all guiding images $\mathbf{I}^G$ by optimizing
\begin{align}
\tilde{\Theta} \coloneqq\argmin_{\Theta} \; \sum_{I_s^G \in \mathbf{I}^G} (F^{\text{image}}_{\Theta}(\mathbf{o}_s) - I_s^G)
\end{align}
\item[\textit{baseline2}: masked mse retraining]
We re-train a new NeRF model using all guiding images $\mathbf{I}^G$ by optimizing:
\begin{align}
\tilde{\Theta} \coloneqq\argmin_{\Theta} \; \sum_{I_s^G \in \mathbf{I}^G} (F^{\text{image}}_{\Theta}(\mathbf{o}_s) - I_s^G) \odot M_s,
\end{align}
where $\Theta$ is randomly initialized.
\end{description}
Both \textit{baseline1} and \textit{baseline2} did not consider depth information during updating the pre-trained NeRF model or re-train a new NeRF model.

We compared the inpainted results of our framework to inpainted results of two baseline methods on LLFF dataset and our custom dataset described in \Cref{sec:custom_dataset}.
For LLFF dataset, we perform qualitative evaluation by applying three methods on each pre-trained model.
For our custom dataset, we perform both qualitative and quantitative evaluations.
For each scene, we trained two separate NeRF models for the \textbf{original} set and the \textbf{removed} set.
We apply three methods to the pre-trained NeRF model using the \textbf{original} set.
We then compared the inpainted result with the image generated by the pre-trained NeRF model using the \textbf{removed} set.

\subsubsection{Results and discussions}
\begin{description}[style=unboxed,leftmargin=0cm]
\item[LLFF dataset]
We show the qualitative comparison between our method and two baseline methods using LLFF dataset in \Cref{fig:llff_qual_comp} and \Cref{fig:consistency}.
In \Cref{fig:llff_qual_comp}, we observed that the depth maps of the inpainted NeRF generated by \textit{baseline1} did not match the inpainted image content.
In \Cref{fig:consistency}, we showed that there are obvious visaul inconsistencies between views in the results generated by \textit{baseline1}. 
To avoid these visual inconsistency, we choose to provide color guidance using only the user-chosen view and let the NeRF model maintain the view consistency by itself.
\textit{Baseline2} recovers visual satisfactory image content without any color guidance inside the masked region. 
However, \textbf{baseline2} still generate results that losses fine structures or synthesize some unnatural patches at complicated regions, which can be oberserved at Horns and Orchids.
\item[Custom dataset] 
We show the qualitative comparison between our method and two baseline methods using our custom dataset in \Cref{fig:teaser} and \Cref{fig:custom_result}. 
In \Cref{fig:custom_result}, we can observe that although \textit{baseline1} can synthesize rgb content closer to ground truth, it still fail to generate correct depth map. 
On the other hand, \textit{baseline2} recovers the content in the masked region guided by the content from different views but still creates noisy and blurry content. 
Our framework generates closer color and depth images to the ground truth rendered results compared to the two baseline methods.
\end{description}
Overall, our inpainting optimization updates a pre-trained NeRF model to obtain correct geometry and preserve visual consistency across views. 

\subsection{Ablation study}
\subsubsection{How important is the depth-guiding loss?}
Introducing the depth-guiding loss ($L_{\text{depth}}$) is one of the major contribution of our framework.
We validate its effectiveness by comparing with the optimization results using color-guiding loss ($L_{\text{color}}$) only, depth-guiding loss ($L_{\text{depth}}$) only, and both losses.

We showed the results in \Cref{fig:depth_ablation}.
We observe that optimizing using $L_{\text{depth}}$ only already leads to correct geometries inside the masked region but introduces color noises outside the masked region.
Our method optimizes both losses and generate correct geometries without color noises.
In \Cref{fig:term_local_comp}, we can also observed that the unwanted object in the
region with high depth variations can be removed by using $L_{\text{depth}}$ only (red box). 
However, using $L_{\text{depth}}$ only
loses color information in the flat region (blue box).
Our method combines these two losses and remove the unwanted object without losing color information in the flat region.


\subsubsection{How important is color-guiding within the masked regions from sampled views?}
In our framework, we only use $L_{\text{color}}^{\text{all}}$ to guide the color reconstruction inside the masked regions. 
We validate this function design by 
adjusting the number of views used to compute $L_{\text{color}}^{\text{all}}$ during the optimization. 

We compared the results of following three settings:
\begin{enumerate}
    \item only \textit{user-chosen} view is used to guide the color inside the masked region, \ie~ $\mathbf{O}^{\text{all}}=\mathbf{o}_u$ and $\mathbf{O}^{\text{out}}=\mathbf{O}$.
    \item \textit{three sampled views} are used to guide the color inside the masked region, \ie~$\mathbf{O}^{\text{all}}=\{\mathbf{o}_i, \mathbf{o}_j, \mathbf{o}_k \}$ where $i,j,k$ are randomly sampled, and $\mathbf{O}^{\text{out}}=\{\mathbf{o}_s| \mathbf{O}-\mathbf{O}^{\text{all}}\}$.
    \item \textit{all sampled views} (\ie~$24$) are used to guide the color inside the masked region, \ie~$\mathbf{O}^{\text{all}}=\mathbf{O}$ and $\mathbf{O}^{\text{out}}=\mathbf{o}_u$.
\end{enumerate}


As shown in \figname~\ref{fig:k_setting}, we observe that more visual inconsistency will be introduced when we use more inpainted images as color guidance. 
Our framework obtains stable results for most of the scene using user-chosen view only; thus, we choose to not to include other inpainted images in the computation of $L_{\text{color}}^{\text{all}}$.

\section{Limitations and future work}
\begin{figure}[h!]
    \includegraphics[width=\linewidth]{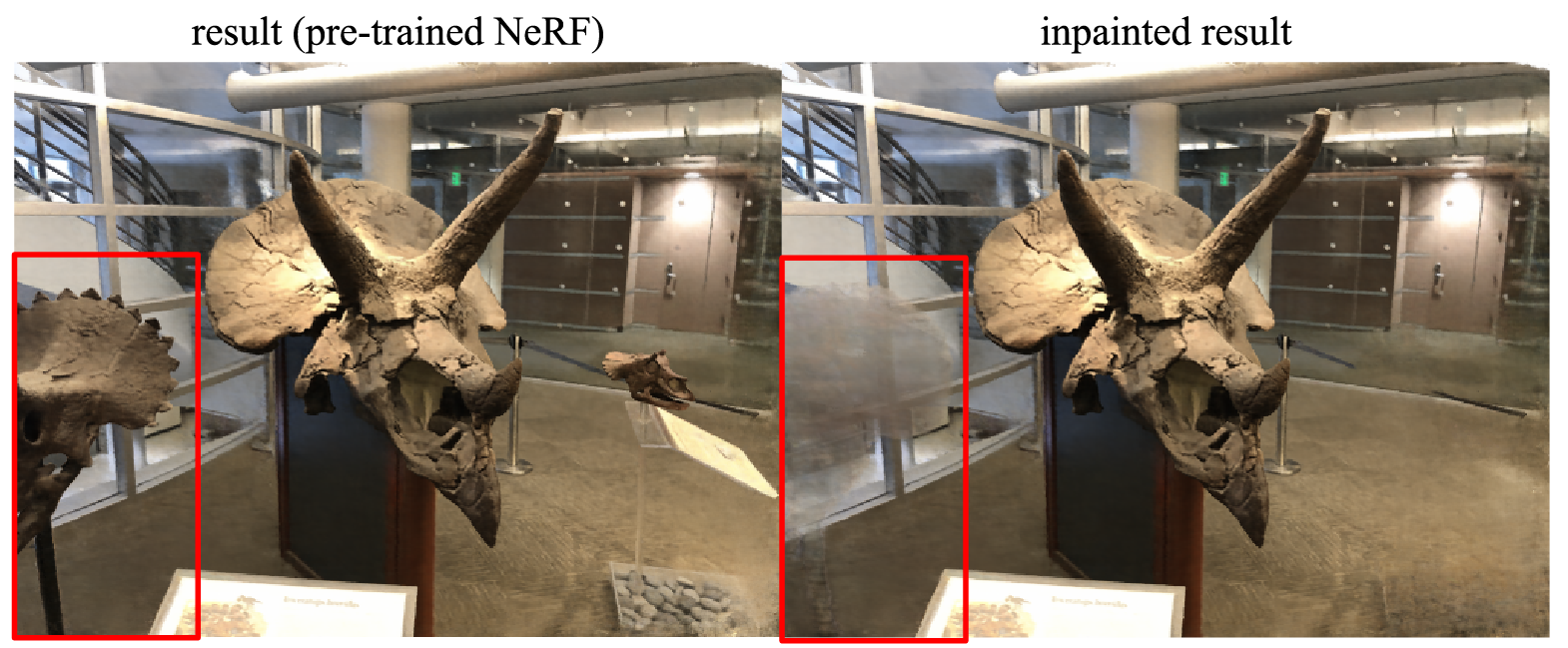}
    \caption{
    Our framework fails to inpaint the mask region in the left region (red box) and introduces artifacts in the optimized results.
    }
    \label{fig:limitation}
\end{figure}

\begin{description}[style=unboxed,leftmargin=0cm]
\item[Fuse color and depth guidance.] 
Our framework leverages existing image inpainting and depth completion method to generate initial guidance materials.
Once there are appearances or geometries artifacts in the initial guidance materials, the optimized NeRF might output undesired or incorrect results.
Meanwhile, our framework shares the same limitations as the image inpainting method we used.
For example, our framework fails to inpaint the image region with high reflectance content (\Cref{fig:limitation}) or with a thin structure.
It is possible to design a fusion method to fuse color and depth guidances from multiple methods. 
\item[Update masks and guidance materials.]
In our current framework, we used fixed masks and guidance materials during the optimization.
However, this is sub-optimal when the unwanted object is occluded in some views.
In the future, we plan to extend our framework to update the masks in every optimization step using 3D volume features extracted from the NeRF model.
We also plan to use a discriminator to constrain the synthesized results and improve the view consistency.
\item[Volume feature for mask transferring]
Our current framework uses the existing video-based object segmentation method to transfer the user-drawn mask.
It is possible to perform mask transferring by conducting 3D volume segmentation using the volume feature extracted from the pre-trained NeRF.

\end{description}
\section{Conclusion}
In this paper, we propose the first framework that enables users to remove unwanted objects or retouch undesired regions in a 3D scene represented by a pre-trained NeRF.
Our framework requires no additional category-specific data and training. 
Instead, we formulated a novel optimization to inpaint the pre-trained NeRF with the generated RGB-D guidances.
We demonstrated our framework handles a variety of scenes well, and we also validate our framework using a custom dataset where ground truth inpainted results are available.
We believe that the custom dataset we proposed and our framework can foster future research on neural radiance field editings.

{\small
\bibliographystyle{ieee_fullname}
\bibliography{paper}
}    
\end{document}